\documentclass[sigconf]{acmart}
\AtBeginDocument{%
  }

\setcopyright{cc}
\setcctype{by}
\copyrightyear{2025}
\acmYear{2025}
\acmConference[MM '25]{Proceedings of the 33rd ACM International Conference on Multimedia}{October 27--31, 2025}{Dublin, Ireland}
\acmBooktitle{Proceedings of the 33rd ACM International Conference on Multimedia (MM '25), October 27--31, 2025, Dublin, Ireland}
\acmDOI{10.1145/3746027.3755592}
\acmISBN{979-8-4007-2035-2/2025/10}

\begin{document}

\title{WMamba: Wavelet-based Mamba for Face Forgery Detection}


\author{Siran Peng}
\email{pengsiran2023@ia.ac.cn}
\affiliation{%
	\institution{MAIS, CASIA; SAI, UCAS}
	\city{Beijing}
	\country{China}
}

\author{Tianshuo Zhang}
\email{tianshuo.zhang@nlpr.ia.ac.cn}
\affiliation{%
	\institution{SAI, UCAS; MAIS, CASIA}
	\city{Beijing}
	\country{China}
}

\author{Li Gao}
\email{gaolids@chinamobile.com}
\affiliation{%
	\institution{CMFT Co., Ltd.}
	\city{Beijing}
	\country{China}
}

\author{Xiangyu Zhu}
\email{xiangyu.zhu@ia.ac.cn}
\affiliation{%
	\institution{MAIS, CASIA; SAI, UCAS}
	\city{Beijing}
	\country{China}
}

\author{Haoyuan Zhang}
\email{zhanghaoyuan2023@ia.ac.cn}
\affiliation{%
	\institution{SAI, UCAS; MAIS, CASIA}
	\city{Beijing}
	\country{China}
}

\author{Kai Pang}
\email{pangkai@pixelall.com}
\affiliation{%
	\institution{Guangzhou Pixel Solutions Co., Ltd.}
	\city{Guangzhou}
	\country{China}
}

\author{Zhen Lei}
\authornote{Corresponding author.}
\email{zhen.lei@ia.ac.cn}
\affiliation{%
	\institution{MAIS, CASIA; SAI, UCAS}
	\city{Beijing}
	\country{China}
}
\affiliation{%
	\institution{CAIR, HKSIS, CAS}
	\city{Hong Kong}
	\country{China}
}
\affiliation{%
	\institution{SCSE, FIE, M.U.S.T}
	\city{Macau}
	\country{China}
}

\renewcommand{\shortauthors}{Siran Peng et al.}

\begin{abstract}
The rapid evolution of deepfake generation technologies necessitates the development of robust face forgery detection algorithms. Recent studies have demonstrated that wavelet analysis can enhance the generalization abilities of forgery detectors. Wavelets effectively capture key facial contours, often slender, fine-grained, and globally distributed, that may conceal subtle forgery artifacts imperceptible in the spatial domain. However, current wavelet-based approaches fail to fully exploit the distinctive properties of wavelet data, resulting in sub-optimal feature extraction and limited performance gains. To address this challenge, we introduce WMamba, a novel wavelet-based feature extractor built upon the Mamba architecture. WMamba maximizes the utility of wavelet information through two key innovations. First, we propose Dynamic Contour Convolution (DCConv), which employs specially crafted deformable kernels to adaptively model slender facial contours. Second, by leveraging the Mamba architecture, our method captures long-range spatial relationships with linear complexity. This efficiency allows for the extraction of fine-grained, globally distributed forgery artifacts from small image patches. Extensive experiments show that WMamba achieves state-of-the-art (SOTA) performance, highlighting its effectiveness in face forgery detection.
\end{abstract}




\maketitle

\begin{figure}[t]
	\begin{center}
		\begin{minipage}{1\linewidth}
			{\includegraphics[width=0.77\linewidth]{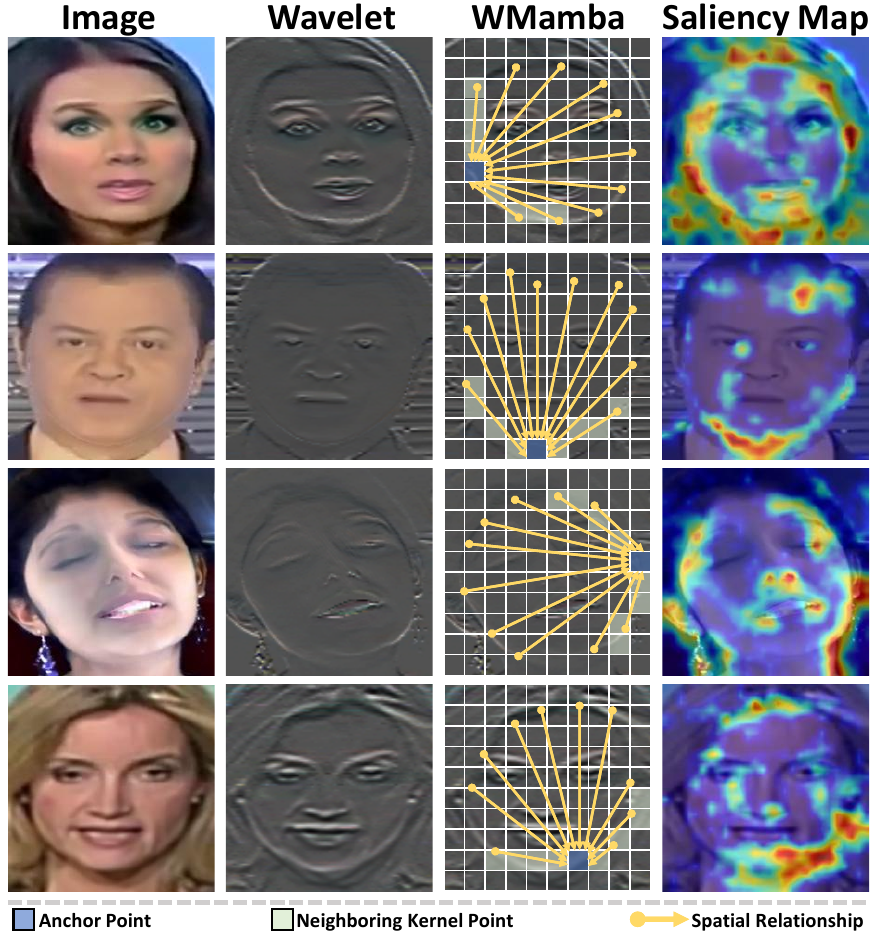}}
			\centering
		\end{minipage}
	\end{center}
	\caption{\textcolor{red}{Column 1}: Example frames from the FaceForensics++ (FF++) dataset \cite{Rossler_2019_ICCV}. \textcolor{red}{Column 2}: Wavelets capture facial contours that are often slender, fine-grained, and globally distributed. \textcolor{red}{Column 3}: WMamba maximizes the potential of wavelet data via two innovations: DCConv for precise modeling of slender facial contours (green squares) and Mamba \cite{gu2023mamba} for extracting fine-grained and globally distributed forgery artifacts (yellow arrows). \textcolor{red}{Column 4}: Saliency maps generated by Grad-CAM \cite{Selvaraju_2017_ICCV} reveal that our model focuses on key facial contours, which are rich in forgery artifacts. \label{hp}}
\end{figure}

\section{Introduction}
Recent advancements in deepfake generation technologies \cite{Nirkin_2019_ICCV,li2019faceshifter,9320155,Hsu_2022_CVPR,Shiohara_2023_ICCV} have garnered significant attention due to their ability to produce highly realistic digital faces. While these technologies offer entertainment value, their potential for misuse, such as facilitating fraud, spreading misinformation, and fabricating fake news, poses serious social risks. As a result, there is an urgent need for the development of robust face forgery detection algorithms.

Detecting facial forgery artifacts directly in the spatial domain presents inherent challenges due to their subtle and often imperceptible nature. Consequently, many approaches turn to frequency analysis techniques to reveal hidden manipulation traces \cite{qian2020thinking,Li_2021_CVPR,tan2024frequency}. Among these techniques, wavelet analysis has gained significant attention for its ability to capture intricate frequency components while preserving essential spatial characteristics \cite{9447758,Liu_2022_ACCV,10.1145/3503161.3547832,10004978}. Specifically, wavelets effectively capture key facial contours that are \textbf{slender} (elongated and narrow), \textbf{fine-grained} (typically one to two pixels wide), and \textbf{globally distributed} (covering large regions across the entire image), as illustrated in Figure~\ref{hp}. These contours often encode subtle but informative forgery clues, making them crucial for manipulation detection, as shown in Figure~\ref{dwt}. However, existing wavelet-based approaches rely on standard convolutions or Transformers \cite{dosovitskiy2020image} for feature extraction, which limits their ability to fully exploit the unique properties of wavelet information.

To tackle this challenge, we introduce WMamba, a wavelet-based feature extractor built upon the Mamba architecture \cite{gu2023mamba}. WMamba fully considers the slender, fine-grained, and globally distributed nature of facial contours, maximizing the potential of wavelet information through two critical innovations. First, inspired by previous studies on deformable convolution \cite{Dai_2017_ICCV,Zhu_2019_CVPR,park2022deformable,Qi_2023_ICCV,10.1145/3664647.3681179}, we present Dynamic Contour Convolution (DCConv), which utilizes meticulously crafted deformable kernels to adaptively capture the \textbf{slender} structures of facial contours. Second, our method leverages the Mamba architecture, a highly effective alternative to traditional Convolutional Neural Networks (CNNs) and Transformers. Rooted in the State Space Model (SSM), Mamba excels at capturing long-range relationships while maintaining linear computational complexity. This efficiency facilitates the use of smaller image patches, enabling the extraction of \textbf{fine-grained} and \textbf{globally distributed} forgery artifacts. In conclusion, our main contributions are as follows:
\begin{itemize}
\item We propose Dynamic Contour Convolution (DCConv), an innovative variant of deformable convolution that employs carefully designed deformable kernels to adaptively and precisely capture the slender structures of facial contours.
\item We demonstrate the effectiveness of the Mamba architecture in detecting facial forgeries. Mamba excels at capturing long-range dependencies while maintaining linear computational complexity, which allows it to extract fine-grained, globally distributed facial forgery clues from small image patches.
\item By combining DCConv with Mamba, the proposed WMamba fully leverages the potential of wavelet information. Extensive experimental results show that WMamba achieves state-of-the-art (SOTA) performance, underscoring its exceptional capabilities in the field of face forgery detection.
\end{itemize}

\section{Related Works \& Motivation}
\subsection{Face Forgery Detection}
According to \cite{pei2024deepfake}, existing studies on face forgery detection can be broadly categorized into four types based on the strategies they employ: spatial-domain, frequency-domain, time-domain, and data-driven approaches.
Spatial-domain methods detect facial forgeries by examining variations in spatial details at the image level, such as color \cite{8803740}, saturation \cite{8803661}, and artifacts \cite{Zhao_2021_CVPR,Cao_2022_CVPR,Shiohara_2022_CVPR,9694644,cui2024forensics}. 
In contrast, time-domain methods analyze inter-frame inconsistencies, leveraging the entire video as input to identify manipulations \cite{Haliassos_2021_CVPR,10.1145/3474085.3475508,gu2022delving,gu2022hierarchical,10054130,10478974,zhang2025learning}.
Frequency-domain methods employ algorithms such as the Fast Fourier Transform (FFT) \cite{tan2024frequency}, Discrete Cosine Transform (DCT) \cite{qian2020thinking,Li_2021_CVPR}, and Discrete Wavelet Transform (DWT) \cite{9447758,Liu_2022_ACCV,10.1145/3503161.3547832,10004978} to convert data from the spatial or time domain into the frequency domain. These techniques are particularly effective in uncovering subtle forgery traces that are often imperceptible in the spatial or time domain.
Finally, data-driven methods focus on optimizing model architectures and refining training strategies to maximize the utility of available data \cite{Zhao_2021_ICCV,hu2022finfer,Huang_2023_CVPR,Guo_2023_CVPR,Zhai_2023_ICCV,Yan_2023_ICCV,Yan_2024_CVPR}. 

\begin{figure}[t]
	\begin{center}
		\begin{minipage}{1\linewidth}
			{\includegraphics[width=1\linewidth]{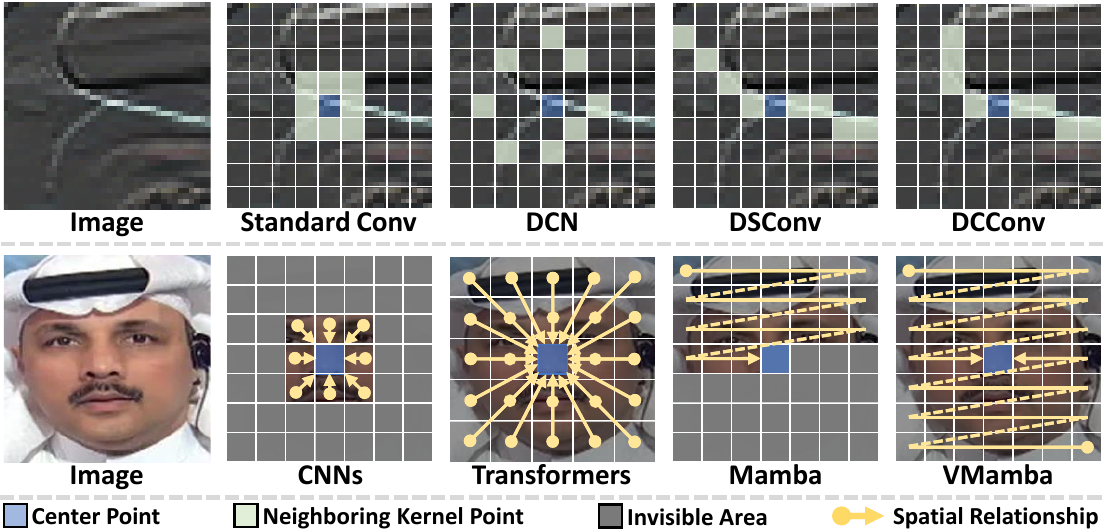}}
			\centering
		\end{minipage}
	\end{center}
	\caption{\textcolor{red}{Row 1}: Graphical comparison of different convolutional paradigms for capturing slender structures. Standard convolutions and DCN struggle with such structures, DSConv offers limited representation, while DCConv demonstrates superior capability. \textcolor{red}{Row 2}: Graphical comparison of global perception capabilities. Only two flattening directions of VMamba are visualized. CNNs lack global perception ability. Transformers capture global context but require splitting the input image into larger patches due to quadratic complexity. Mamba exhibits partial global perception, while VMamba demonstrates enhanced global perception capability. \label{comprr}}
\end{figure}

\begin{figure*}[t]
	\begin{center}
		\begin{minipage}{1\linewidth}
			{\includegraphics[width=0.97\linewidth]{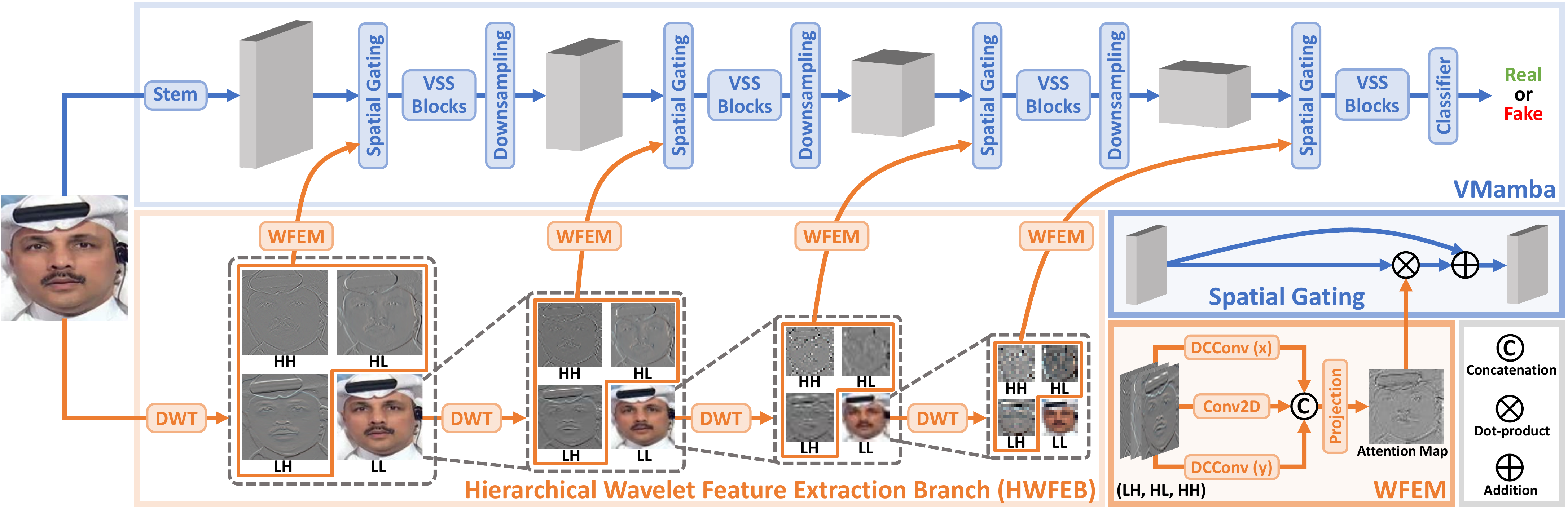}}
			\centering
		\end{minipage}
	\end{center}
	\caption{Overview of WMamba. The architecture comprises two main components: HWFEB and VMamba. HWFEB employs multi-level DWT to capture wavelet representations, WFEMs to generate spatial attention maps, and spatial gating mechanisms to integrate these maps into VMamba. VMamba then extracts wavelet-enhanced forgery cues and performs classification. \label{pipeline}}
\end{figure*}

\subsection{Deformable Convolution}
\label{s22}
Standard convolution operations rely on fixed kernel shapes, which hinders their adaptability to objects exhibiting diverse structures and orientations. To address this limitation, the Deformable Convolutional Network (DCN) \cite{Dai_2017_ICCV} introduces learnable offsets that dynamically adjust the kernel shape, allowing for better alignment with the target's geometry. While this innovation enhances the network's ability to model complex geometric structures, DCN faces difficulties when applied to slender shapes. To overcome this challenge, \cite{Qi_2023_ICCV} proposes the Dynamic Snake Convolution (DSConv), which simulates slender structures by imposing constraints on both the initial kernel shape and the offset learning process. In 2D scenarios, for instance, it initializes with a 1D kernel oriented along a fixed coordinate axis (either x or y). The kernel points form an equidistant sequence along this primary axis. DSConv then iteratively calculates the perpendicular offsets for each kernel point, with a key constraint: the offset difference between neighboring points cannot exceed one. This design enables the representation of slender structures with smooth and continuous topologies. However, the reliance on a predefined axis restricts DSConv's ability to capture objects with arbitrary directions. In contrast, the proposed DCConv dynamically learns the optimal axis orientation, allowing for the adaptation to a wider variety of slender structures. These convolutional paradigms are illustrated in Row 1 of Figure~\ref{comprr}.

\subsection{Mamba}
\label{s23}
Deep learning-driven computer vision primarily relies on CNNs and Transformers for feature extraction. While CNNs are efficient, their limited receptive fields hinder their ability to capture global context. In contrast, Transformers excel at modeling long-range relationships but suffer from quadratic computational complexity. To address this issue, \textbf{Transformer-based methods generally divide input images into large patches, which results in the loss of fine-grained spatial details.} Recently, the Mamba architecture \cite{gu2023mamba}, based on the SSM, has emerged as a promising alternative by achieving global perception with linear computational complexity. However, Mamba was originally designed for 1D tasks with inherent directional structures. Extending it directly to 2D vision tasks, where such directional patterns are often absent, can lead to incomplete global perception. To tackle this limitation, Vision Mamba \cite{10.5555/3692070.3694654} introduces a bidirectional flattening approach that flattens spatial feature maps along both positive and negative directions, resulting in a more comprehensive global perception. Building on this foundation, VMamba \cite{liu2024vmamba} proposes a four-directional flattening technique, enabling the discovery of richer spatial relationships. These methods are visually illustrated in Row 2 of Figure~\ref{comprr}. 

\subsection{Motivation}
\begin{figure*}[t]
	\begin{center}
		\begin{minipage}{1\linewidth}
			{\includegraphics[width=0.9\linewidth]{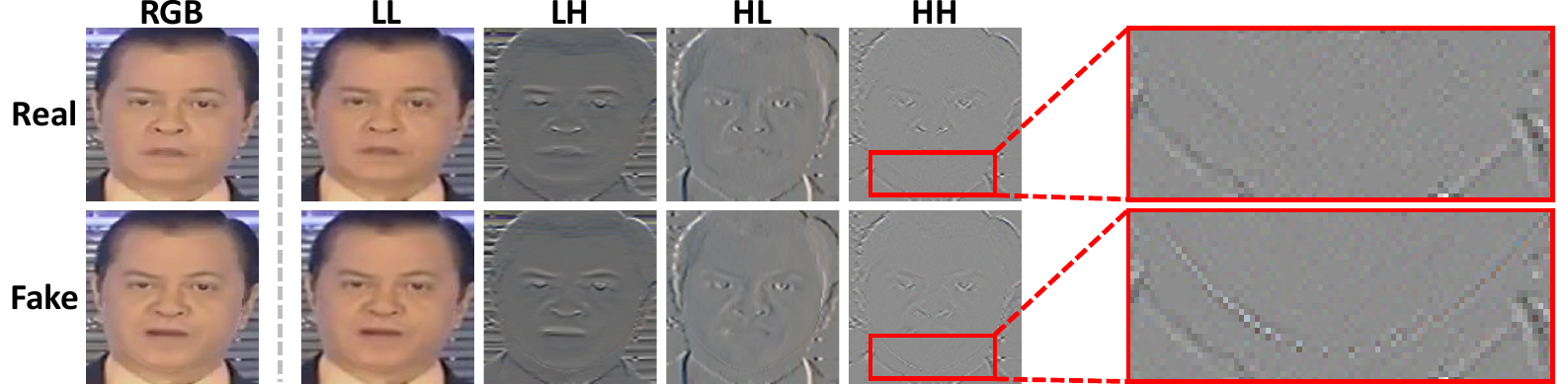}}
			\centering
		\end{minipage}
	\end{center}
	\caption{Visualization of different frequency sub-bands from the DWT. The LH, HL, and HH sub-bands capture high-frequency details in various orientations, while the LL sub-band represents a low-resolution approximation of the original RGB image. Highlighted within the red boxes, the high-frequency sub-bands reveal critical forgery traces that are otherwise less apparent.
		\label{dwt}}
\end{figure*}

Wavelets effectively capture key facial contours, which tend to be slender, fine-grained, and globally distributed, encoding crucial forgery clues essential for detecting manipulations, as demonstrated in Figures~\ref{hp} and \ref{dwt}. However, existing wavelet-based detectors primarily rely on standard convolutions or Transformers for feature extraction, both of which exhibit notable limitations. Standard convolutions struggle to capture slender structures and global context, while Transformers, although capable of modeling long-range dependencies, fail to preserve fine-grained details and delicate geometries. To address the challenge of simulating slender structures, we draw inspiration from DSConv, which constrains its initial kernel shape and offset learning process to better align with such geometries. However, DSConv's reliance on a fixed axis significantly limits its flexibility. Therefore, we propose \textbf{DCConv}, a novel approach that dynamically learns the optimal axis orientation, enabling robust modeling of slender structures along arbitrary directions. For capturing both fine-grained and globally distributed features, we leverage the \textbf{Mamba} architecture. Mamba achieves global perception with linear complexity, making it far more efficient than Transformers. This efficiency allows for the use of smaller image patches, facilitating the extraction of fine-grained and globally distributed characteristics. By integrating DCConv with Mamba, the proposed WMamba fully exploits the unique properties of wavelet representations, offering a more interpretable and effective solution compared to previous wavelet-based detectors.

\section{Methodology}
\subsection{Overview}
The proposed WMamba network architecture, depicted in Figure~\ref{pipeline}, consists of two primary components: the Hierarchical Wavelet Feature Extraction Branch (HWFEB) and the VMamba model. For a given RGB input image, the HWFEB first applies a multi-level DWT to extract wavelet representations across various scales. Subsequently, these representations are processed by distinct Wavelet Feature Extraction Modules (WFEMs), which primarily leverage DCConv to generate spatial attention maps. Next, these attention maps are seamlessly integrated into different stages of the VMamba model using spatial gating mechanisms. The model ultimately outputs two probabilities, classifying the input image as either real or fake. Since this paper focuses on the network architecture, we adopt the standard Cross-Entropy (CE) Loss function for training. The remainder of this section is organized as follows: Section~\ref{wmamba} provides a comprehensive description of the WMamba pipeline, while Section~\ref{dcconv} offers an in-depth explanation of DCConv.

\subsection{WMamba}
\label{wmamba}
In this section, we will introduce the two key components of WMamba: (1) the HWFEB, which includes a multi-level DWT, multiple WFEMs, and spatial gating mechanisms, as well as (2) the VMamba model.

\subsubsection{HWFEB} 
We begin by utilizing a \textbf{multi-level Haar DWT} to generate wavelet representations at different scales. For an input image $I\in\mathbb{R}^{H\times W\times 3}$, where $H$ and $W$ denote the height and width, the Haar DWT operates by convolving $I$ with four distinct filters: $f_{\rm{LL}}$, $f_{\rm{LH}}$, $f_{\rm{HL}}$, and $f_{\rm{HH}}$. These filters correspond to the Low-Low, Low-High, High-Low, and High-High frequency sub-bands, producing the respective wavelet sub-band outputs: LL, LH, HL, and HH, each of size $\frac{H}{2}\times \frac{W}{2}\times 3$. This process can be expressed as follows:
\begin{equation}\label{E1}
	\setlength{\arraycolsep}{1pt}
	\renewcommand{\arraystretch}{1.25}
	\begin{aligned}
		{\rm{LL}}, {\rm{LH}}, {\rm{HL}}, {\rm{HH}} &= (f_{\rm{LL}}, f_{\rm{LH}}, f_{\rm{HL}}, f_{\rm{HH}})\circledast I, \\
		f_{\rm{LL}}, f_{\rm{LH}}, f_{\rm{HL}}, f_{\rm{HH}} &= \frac{1}{2}
		\begin{bmatrix}
			1 & 1 \\
			1 & 1 \\
		\end{bmatrix}, \frac{1}{2}
		\begin{bmatrix}
			1 & 1 \\
			-1 & -1 \\
		\end{bmatrix}, \frac{1}{2}
		\begin{bmatrix}
			1  & -1  \\
			1  & -1  \\
		\end{bmatrix}, \frac{1}{2}
		\begin{bmatrix}
			1  & -1  \\
			-1  & 1  \\
		\end{bmatrix}.
	\end{aligned}
	\renewcommand{\arraystretch}{1}
\end{equation}
Here, $\circledast$ represents the convolution operation. As illustrated in Figure~\ref{dwt}, the LH, HL, and HH sub-bands capture high-frequency details, such as edges and textures, from different orientations. In contrast, the LL sub-band essentially serves as a low-resolution approximation of the input image and can be recursively decomposed. This recursive process forms a multi-level DWT, which enables the generation of wavelet representations across various scales. 

Subsequently, we leverage \textbf{WFEMs} to generate spatial attention maps from the wavelet sub-bands LH, HL, and HH across multiple scales. Each WFEM extracts forgery-related features using three parallel convolutional layers: two DCConv layers (initialized along the x-axis and y-axis, respectively) and one standard 2D convolutional layer. The DCConv layers specialize in capturing the slender structures of facial contours, while the standard convolutional layer focuses on learning broader, more generalized patterns. The resulting feature maps from these layers are then concatenated and projected to create the desired spatial attention map.

Finally, we incorporate spatial attention maps into multiple stages of the VMamba model through \textbf{spatial gating mechanisms}. At each stage, a dot-product operation is performed to combine the attention map with the model's feature map, enabling spatially adaptive weighting. This process effectively highlights critical features essential for identifying forgeries. In addition, we employ skip connections to enhance model stability and accelerate convergence.

HWFEB distinguishes itself from MWFEB \cite{10.1145/3503161.3547832} in two key aspects: (1) HWFEB exclusively uses high-frequency wavelet sub-bands, while MWFEB incorporates all wavelet components. As shown in Table~\ref{HWFEB}, our selective strategy achieves superior performance. (2) HWFEB introduces the novel DCConv to generate spatial attention maps, whereas MWFEB relies solely on standard convolutions.

\subsubsection{VMamba} 
The Mamba architecture is based on the \textbf{SSM} \cite{HAMILTON19943039}, a foundational framework widely used in fields such as control theory, signal processing, and econometrics. The SSM transforms a 1D input signal $x(t)\in\mathbb{R}$ into a 1D output signal $y(t)\in\mathbb{R}$ through the continuous-time evolution of hidden states $h(t)\in\mathbb{R}^{N}$, where $N$ represents the state dimension. This dynamic system is defined by the following set of Ordinary Differential Equations (ODEs):
\begin{equation}\label{E6}
	\begin{aligned}
		h'(t)&={A}h(t)+{B}x(t), \\
		y(t)&={C}h(t)+{D}x(t).
	\end{aligned}
\end{equation} 
Here, ${A}\in \mathbb{R}^{{N\times N}}$ is the state matrix governing the system's temporal evolution. ${B}\in \mathbb{R}^{{N\times 1}}$, ${C}\in \mathbb{R}^{{1\times N}}$, and ${D}\in \mathbb{R}^{{1\times 1}}$ are input, output, and feedthrough projection parameters, respectively. The system exhibits global memory characteristics, as evidenced by Equation~\ref{E6}, where each output depends on the entire history of past inputs through recursive state propagation. 
When implementing the SSM in deep learning applications, discretization is required to transform the continuous-time system into its discrete-time equivalent. This process introduces a timescale parameter $\Delta \in \mathbb{R}$, which maps the continuous-time parameters $A$ and $B$ to their discrete-time counterparts $\overline{A}$ and $\overline{B}$. By employing the Zero-Order Hold (ZOH) algorithm, the discrete-time parameters can be calculated as follows:
\begin{equation}\label{E7}
	\begin{aligned}
		{\overline{A}}&=e^{\Delta A}, \\
		{\overline{B}}&={({\Delta}{A})}^{-1}(e^{\Delta A}-{{E}})\cdot {\Delta}{B}\approx{\Delta}{B}.
	\end{aligned}
\end{equation} 
Here, $E\in\mathbb{R}^{N\times N}$ represents the identity matrix. Then, the discrete form of Equation~\ref{E6} can be expressed as follows:
\begin{equation}\label{E8}
	\begin{aligned}
		h_t&={\overline{A}}h_{t-1}+{\overline{B}}x_{t}, \\
		y_t&={C}h_t+{D}x_{t}.
	\end{aligned}
\end{equation} 
In practice, $x_t$ is a feature vector with multiple components, each of which is processed independently according to Equation~\ref{E8}.

\begin{figure}[t]
	\begin{center}
		\begin{minipage}{1\linewidth}
			{\includegraphics[width=0.99\linewidth]{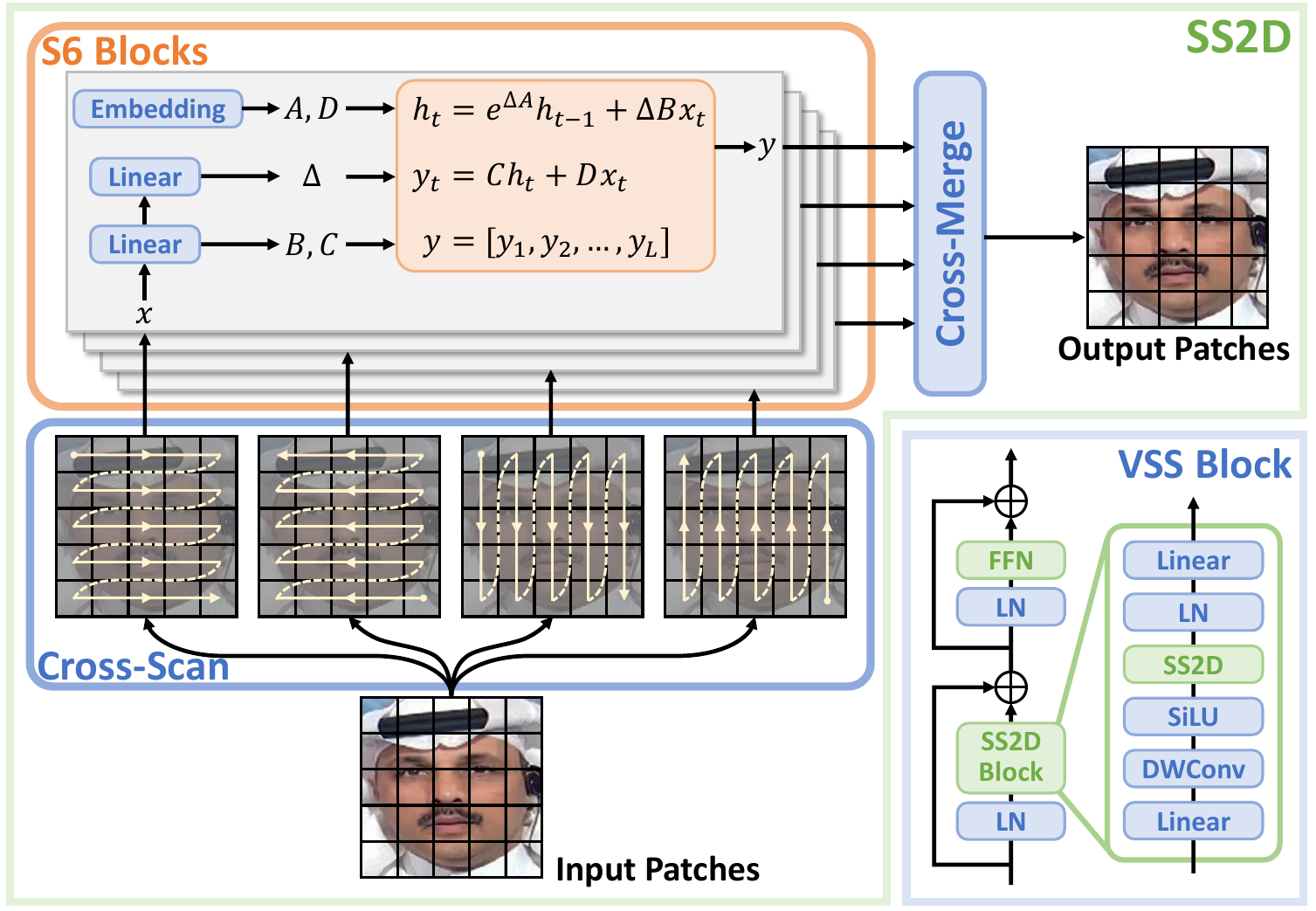}}
			\centering
		\end{minipage}
	\end{center}
	\caption{Schematic diagram of the VSS block, with the SS2D mechanism at its core. This mechanism flattens input image patches along four principle directions, facilitating comprehensive global perception. $L$ denotes the number of patches. \label{vss}}
\end{figure}

The \textbf{VMamba} model \cite{liu2024vmamba}, a 2D adaptation of Mamba, utilizes hierarchical Visual State Space (VSS) blocks to extract rich forgery signatures from image patches across multiple spatial resolutions. As illustrated in Figure~\ref{vss}, each VSS block consists of two key components: a Selective-Scan 2D (SS2D) block and a Feed Forward Network (FFN). The SS2D block performs channel projection operations and employs an SS2D mechanism to capture fine-grained, globally distributed forgery clues from input image patches. Following this, the FFN extracts channel-specific facial features, further enhancing the model's ability to learn and identify forgery artifacts.

The SS2D mechanism comprises three core components: a cross-scan module, a stack of four S6 blocks, and a cross-merge module. The cross-scan module scans input image patches along four principal directions (horizontal, vertical, and both diagonals), creating flattened sequences that effectively capture multi-directional spatial relationships. Central to the SS2D is the S6 block, a pivotal innovation introduced by Mamba \cite{gu2023mamba}, which integrates a selective-scan mechanism to enhance Equation~\ref{E8}. This mechanism dynamically learns projection and timescale parameters from the input data, thereby boosting the model's flexibility and expressiveness. Finally, the cross-merge module reconstructs the original spatial structure by unflattening the processed sequences and merging them back into a cohesive output image through fusion operations.

\begin{figure}[t]
	\begin{center}
		\begin{minipage}{1\linewidth}
			{\includegraphics[width=0.94\linewidth]{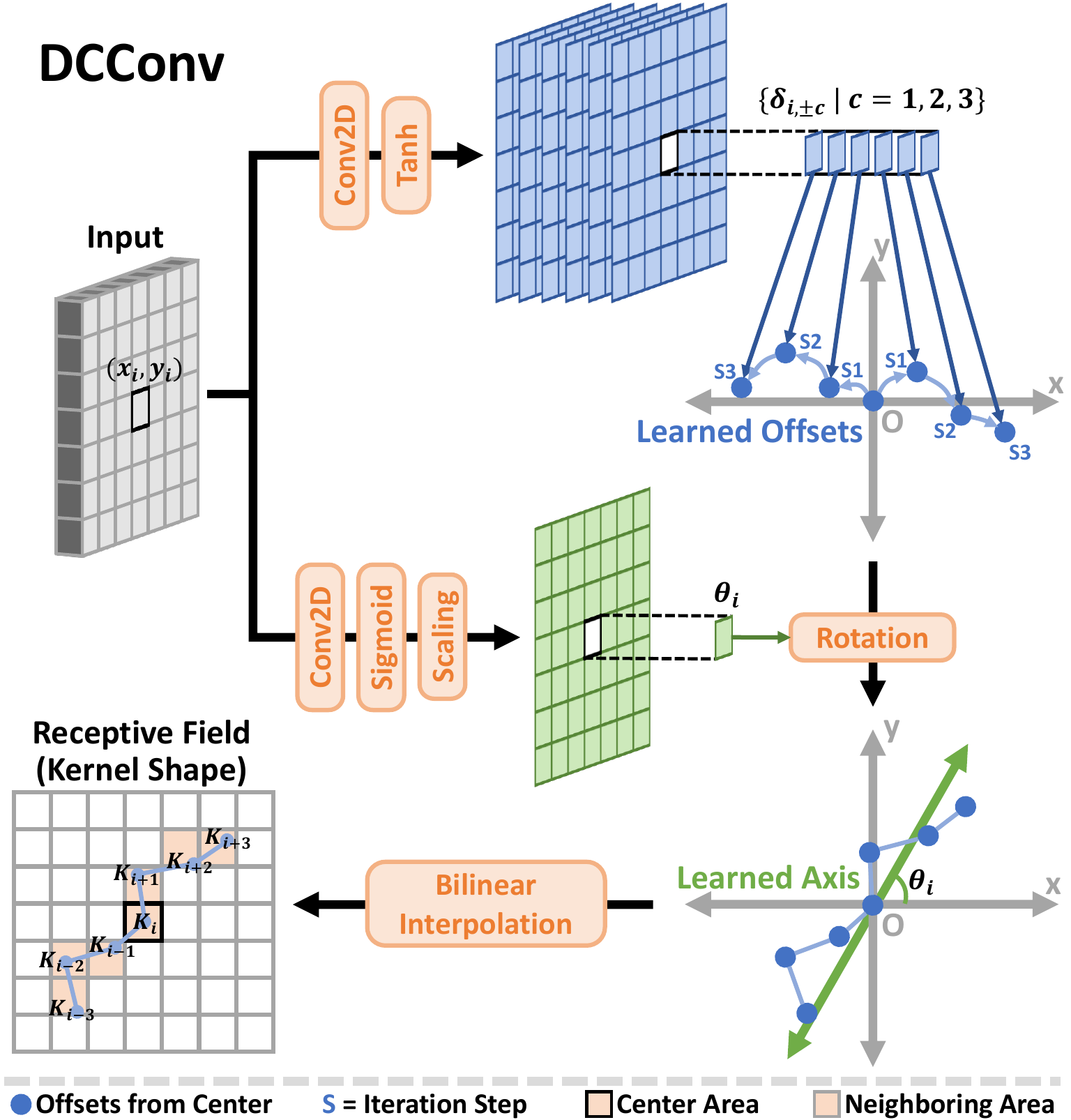}}
			\centering
		\end{minipage}
	\end{center}
	\caption{Schematic diagram of the proposed DCConv, initialized along the x-axis. Our method predicts both the offsets and axis orientations simultaneously, allowing for the effective representation of slender structures aligned in arbitrary directions. For illustration, the kernel length $k$ is set to 7. \label{dcconv_fig}}
\end{figure}

\begin{table*}[t]	
	\centering\renewcommand\arraystretch{1.2}\setlength{\tabcolsep}{11pt}
	\belowrulesep=0pt\aboverulesep=0pt
	\caption{Cross-dataset evaluation of representative face forgery detection methods on the CDF, DFDC, DFDCP, and FFIW datasets. Methods marked with the symbol $\ast$ are reproduced using the official codes, while results for other methods are taken directly from their respective papers. The best performance is highlighted in \textbf{BOLD}, and the second best is \underline{underlined}. Notably, WMamba achieves SOTA performance across all datasets, demonstrating exceptional generalization capability. \label{cross-data}}
	\begin{tabular}{c|c|c|cc|cccc}
		\toprule
		\multirow{2}{*}{Method} & 
		\multirow{2}{*}{Venue} & 
		\multirow{2}{*}{Input Type} &
		\multicolumn{2}{c|}{Training Set} & \multicolumn{4}{c}{Test Set AUC (\%)}\\
		\cmidrule(lr){4-5}\cmidrule(lr){6-9}
		&\multicolumn{1}{c|}{} 
		&\multicolumn{1}{c|}{} 
		&\multicolumn{1}{c}{Real} 
		&\multicolumn{1}{c|}{Fake} 
		&\multicolumn{1}{c}{CDF} 
		&\multicolumn{1}{c}{DFDC} 
		&\multicolumn{1}{c}{DFDCP}
		&\multicolumn{1}{c}{FFIW} \\
		\midrule
		F\textsuperscript{3}-Net\textsuperscript{$\ast$} \cite{qian2020thinking} & ECCV 2020 & Frame & \checkmark & \checkmark & 77.92 & 67.35 & 73.54 & 70.11 \\
		LTW\textsuperscript{$\ast$} \cite{sun2021domain} & AAAI 2021 & Frame & \checkmark & \checkmark & 77.14 & 69.00 & 74.58 & 76.63 \\
		PCL+I2G \cite{Zhao_2021_ICCV} & ICCV 2021 & Frame & \checkmark &  & 90.03 & 67.52 & 74.37 & - \\
		DCL \cite{sun2022dual} & AAAI 2022 & Frame & \checkmark & \checkmark & 82.30 & - & 76.71 & 71.14 \\
		SBI \cite{Shiohara_2022_CVPR} & CVPR 2022 & Frame & \checkmark &  & 93.18 & 72.42 & 86.15 & {84.83} \\
		F\textsuperscript{2}Trans \cite{10004978} & TIFS 2023 & Frame & \checkmark & \checkmark & 89.87 & - & 76.15 & - \\
		SeeABLE \cite{Larue_2023_ICCV} & ICCV 2023 & Frame & \checkmark &  & 87.30 & 75.90 & 86.30 & - \\
		AUNet \cite{Bai_2023_CVPR} & CVPR 2023 & Frame & \checkmark &  & 92.77 & 73.82 & 86.16 & 81.45 \\
		LAA-Net \cite{Nguyen_2024_CVPR} & CVPR 2024 & Frame & \checkmark &  & 95.40 & - & 86.94 & - \\
		RAE \cite{10.1007/978-3-031-72943-0_23} & ECCV 2024 & Frame & \checkmark &  & \underline{95.50} & {80.20} & \underline{89.50} & - \\
		FreqBlender \cite{zhou2024freqblender} & NeurIPS 2024 & Frame & \checkmark &  & 94.59 & 74.59 & 87.56 & \underline{86.14} \\
		UDD \cite{fu2025exploring} & AAAI 2025 & Frame & \checkmark & \checkmark & 93.10 & \underline{81.20} & 88.10 & - \\
		LESB \cite{Soltandoost_2025_WACV} & WACVW 2025 & Frame & \checkmark &  & 93.13 & 71.98 & - & 83.01 \\
		\midrule
		FTCN\textsuperscript{$\ast$} \cite{Zheng_2021_ICCV} & ICCV 2021 & Video & \checkmark & \checkmark & 86.90 & 71.00 & 74.00 & 74.47 \\
		RealForensics \cite{Haliassos_2022_CVPR} & CVPR 2022 & Video & \checkmark & \checkmark & 86.90 & - & 75.90 & - \\
		TALL \cite{Xu_2023_ICCV} & ICCV 2023 & Video & \checkmark & \checkmark & 90.79 & - & 76.78 & - \\
		TALL++ \cite{xu2024towards} & IJCV 2024 & Video & \checkmark & \checkmark & 91.96 & - & 78.51 & - \\
		NACO \cite{zhang2025learning} & ECCV 2024 & Video & \checkmark & \checkmark & 89.50 & - & 76.70 & - \\
		\midrule
		WMamba (Ours) & - & Frame & \checkmark &  & \textbf{96.29} & \textbf{82.97} & \textbf{89.62} & \textbf{86.59} \\      
		\bottomrule
	\end{tabular}
\end{table*}

\subsection{DCConv}
\label{dcconv}
Building on the work of DSConv \cite{Qi_2023_ICCV}, we introduce DCConv, which incorporates a learnable coordinate axis to effectively capture slender structures oriented along arbitrary directions, as demonstrated in Figure~\ref{dcconv_fig}. Specifically, DCConv begins with a 1D convolutional kernel initially aligned along a predefined coordinate axis (either the x-axis or y-axis). For each input feature map, DCConv employs two standard 2D convolutional layers to independently generate pixel-wise offsets and rotation angles. The predicted offsets are bounded within the interval $[-1, 1]$ using a Tanh activation function, while the rotation angles are constrained to the range $[0, \frac{\pi}{2}]$ through a Sigmoid activation function followed by a scaling operation. Let's consider a 1D kernel of odd length $k$, centered at the $i$-th pixel with coordinates $(x_i,y_i)$. The positions of the kernel points are denoted by $\{K_{i\pm c}=(x_{i\pm c}, y_{i\pm c})\ \vert \ c=0,1,...,\frac{k-1}{2}\}$. The corresponding learned offsets for these kernel points are represented by $\{\delta_{i,\pm c}\ \vert \ c=0,1,...,\frac{k-1}{2}\}$, and the predicted rotation angle at the $i$-th pixel is denoted by $\theta_i$. When the 1D kernel is initialized along the x-axis, the adjusted positions of the kernel points are computed as:
\begin{equation}\label{E4}
	K_{i\pm c} = (x_i, y_i) + (\pm c, \sum\limits_{j=0}^{c}\delta_{i,\pm j})\cdot 
	\begin{bmatrix}
		\cos \theta_i  & \sin \theta_i  \\
		-\sin \theta_i  & \cos \theta_i  \\
	\end{bmatrix}.
\end{equation}
In practice, the value of $\delta_{i,0}$ is set to $0$, and the kernel positions are calculated iteratively in a step-by-step manner. When the 1D kernel is initialized along the y-axis, Equation \ref{E4} can be reformulated as:
\begin{equation}\label{E5}
	K_{i\pm c} = (x_i, y_i) + (\sum\limits_{j=0}^{c}\delta_{i,\pm j}, \pm c)\cdot 
	\begin{bmatrix}
		\cos \theta_i  & \sin \theta_i  \\
		-\sin \theta_i  & \cos \theta_i  \\
	\end{bmatrix}.
\end{equation}

The combination of the Tanh activation function and the iterative calculation process ensures that the offset differences between adjacent kernel points remain consistently below one, facilitating the effective extraction of slender structures with smooth and continuous topologies. \textbf{Furthermore, the learned rotation angles enable DCConv to theoretically model slender structures oriented in any direction.} This flexibility allows DCConv to capture a wider range of slender geometries compared to DSConv, which can be regarded as a special case of DCConv when $\theta_i = 0$. A visual comparison of these two convolutional paradigms are provided in Appendix~\ref{appendixa}. Since the offsets are typically fractional, bilinear interpolation is used to compute the receptive field values (\emph{i.e.}, to sample the input feature map) during the convolution operation, as described in \cite{Dai_2017_ICCV}.

\section{Experiments}
\subsection{Setup}
\subsubsection{Datasets}
For training, we utilize the widely recognized FaceForensics++ ({FF++}) \cite{Rossler_2019_ICCV} benchmark dataset, which comprises 1,000 authentic face videos and 4,000 manipulated videos generated using four distinct forgery techniques: DeepFakes ({DF}), Face2Face ({F2F}) \cite{Thies_2016_CVPR}, FaceSwap ({FS}), and NeuralTextures ({NT}) \cite{thies2019deferred}. 
Notably, this study leverages the SBI framework \cite{Shiohara_2022_CVPR}, enabling our model to be trained exclusively on real face videos. This approach supports cross-manipulation evaluation across all four forgery types in FF++. To assess cross-dataset generalizability, we test our model on four popular datasets: Celeb-DeepFake-v2 ({CDF}) \cite{Li_2020_CVPR}, which employs advanced deepfake methods on YouTube celebrity content; the DeepFake Detection Challenge ({DFDC}) \cite{dolhansky2020deepfake} and its Preview version ({DFDCP}) \cite{dolhansky2019dee}, which feature videos with various perturbations including compression, downsampling, and noise; and FFIW-10K ({FFIW}) \cite{Zhou_2021_CVPR}, which adds complexity with multi-person scenarios. 

\subsubsection{Frame-Level Baselines}
We evaluate our approach against thirteen SOTA frame-level baselines for face forgery detection, each employing distinct strategies. F\textsuperscript{3}-Net \cite{qian2020thinking} leverages frequency-aware manipulation clues, while LTW \cite{sun2021domain} emphasizes domain-general detection. PCL+I2G \cite{Zhao_2021_ICCV} focuses on source feature inconsistencies, and DCL \cite{sun2022dual} employs multi-granular contrastive learning. SBI \cite{Shiohara_2022_CVPR} enhances training by generating synthetic fake faces, and F\textsuperscript{2}Trans \cite{10004978} integrates spatial and frequency domain forgery traces. SeeABLE \cite{Larue_2023_ICCV} reformulates forgery detection as an Out-Of-Distribution (OOD) problem, while AUNet \cite{Bai_2023_CVPR} explores facial Action Unit (AU) regions for more nuanced analysis. LAA-Net \cite{Nguyen_2024_CVPR} incorporates heatmap-guided self-consistency attention, and RAE \cite{10.1007/978-3-031-72943-0_23} aims to recover real facial appearances from perturbations. FreqBlender \cite{zhou2024freqblender} synthesizes pseudo-fake faces by blending frequency knowledge, UDD \cite{fu2025exploring} addresses position and content biases with a dual-branch architecture, and LESB \cite{Soltandoost_2025_WACV} introduces a Local Feature Discovery (LFD) technique to produce self-blended samples.

\subsubsection{Video-Level Baselines}
We further compare our method with five video-level face forgery detection approaches. These include {FTCN} \cite{Zheng_2021_ICCV}, which exploits temporal coherence patterns; {RealForensics} \cite{Haliassos_2022_CVPR}, which employs a two-stage framework trained on natural talking face datasets; {TALL} \cite{Xu_2023_ICCV}, which preserves spatio-temporal dependencies through layout transformation; the enhanced version {TALL++} \cite{xu2024towards}, which incorporates a Graph Reasoning Block (GRB) and Semantic Consistency (SC) loss; and {NACO} \cite{zhang2025learning}, which learns natural consistency representations from authentic face videos.

\begin{table}[t]	
	\centering\renewcommand\arraystretch{1.2}\setlength{\tabcolsep}{6.8pt}
	\belowrulesep=0pt\aboverulesep=0pt
	\caption{Cross-manipulation evaluation of methods trained only on real face videos from FF++. Notably, our method delivers the best results across all tested manipulations. \label{cross-mani}}
	\begin{tabular}{c|cccc|c}
		\toprule
		\multirow{2}{*}{Method} & 
		\multicolumn{5}{c}{Test Set AUC (\%)} \\
		\cmidrule(lr){2-6}
		&\multicolumn{1}{c}{DF} 
		&\multicolumn{1}{c}{F2F} 
		&\multicolumn{1}{c}{FS}
		&\multicolumn{1}{c|}{NT}
		&\multicolumn{1}{c}{FF++} \\
		\midrule
		PCL+I2G \cite{Zhao_2021_ICCV} & \textbf{100} & 98.97 & 99.86 & 97.63 & 99.11 \\
		SBI \cite{Shiohara_2022_CVPR} & \underline{99.99} & \underline{99.88} & \underline{99.91} & \underline{98.79} & \underline{99.64} \\
		SeeABLE \cite{Larue_2023_ICCV} & 99.20 & 98.80 & 99.10 & 96.90 & 98.50 \\
		AUNet \cite{Bai_2023_CVPR} & 99.98 & 99.60 & 99.89 & 98.38 & 99.46 \\
		RAE \cite{10.1007/978-3-031-72943-0_23} & 99.60 & 99.10 & 99.20 & 97.60 & 98.90 \\
		\midrule
		WMamba & \textbf{100} & \textbf{99.98} & \textbf{99.94} & \textbf{98.88} & \textbf{99.70}\\      
		\bottomrule
	\end{tabular}
\end{table}

\begin{table}[t]	
	\centering\renewcommand\arraystretch{1.2}\setlength{\tabcolsep}{6.4pt}
	\belowrulesep=0pt\aboverulesep=0pt
	\caption{Ablation study results on the structure of HWFEB. \label{HWFEB}}
	\begin{tabular}{c|cccc}
		\toprule
		\multirow{2}{*}{Method} & 
		\multicolumn{4}{c}{Test Set AUC (\%)} \\
		\cmidrule(lr){2-5}
		&\multicolumn{1}{c}{CDF} 
		&\multicolumn{1}{c}{DFDC} 
		&\multicolumn{1}{c}{DFDCP}
		&\multicolumn{1}{c}{FFIW} \\
		\midrule
		w/ LL & \underline{95.61} & 80.25 & 87.47 & 85.68  \\
		w/o Hierarchical & {94.76} & \underline{81.79} & \underline{87.88} & {85.03} \\
		\midrule
		w/ Addition & 94.70 & 77.57 & 87.13 & \underline{86.34} \\
		w/ Concatenation & 84.54 & 69.19 & 78.22 & 65.05 \\
		w/o Skip Connection & 95.14 & 78.51 & 85.75 & 84.61 \\
		\midrule
		The Proposed & \textbf{96.29} & \textbf{82.97} & \textbf{89.62} & \textbf{86.59}\\ 
		\bottomrule
	\end{tabular}
\end{table}

\subsubsection{Evaluation Metric}
We evaluate the detection performance of various methods using the Area Under the Receiver Operating Characteristic Curve (AUC), a standard metric in face forgery detection. For frame-level methods, we report video-level results, which average the predictions across all frames within each video, allowing for a fair and direct comparison with video-level approaches.

\subsubsection{Implementation Details}
This work leverages the synthetic data generation framework, preprocessing techniques, and data augmentation strategies introduced by SBI \cite{Shiohara_2022_CVPR}. For further details, please refer to Appendix~\ref{appendixb}. We adopt VMamba-S \cite{liu2024vmamba} as our backbone network, which consists of 2, 2, 15, and 2 VSS blocks across its four stages. The network is pre-trained on the ImageNet-1K dataset \cite{5206848}, providing a strong initialization. In addition, we configure DCConv with a kernel length of 9. The WMamba model is trained using the AdamW optimizer \cite{loshchilov2018decoupled} with a batch size of 64 and an initial learning rate of 5e-5. Training spans 200 epochs, during which the learning rate is linearly decayed starting from the 100th epoch to facilitate smoother convergence. All experiments are implemented in PyTorch, and training WMamba under these configurations requires approximately 30 GB of GPU memory.

\subsection{Results}
\subsubsection{Cross-Dataset Evaluation} 
Cross-dataset evaluation is a crucial testing protocol for face forgery detection, as it measures a model's ability to generalize to unseen data. In Table~\ref{cross-data}, we compare our method against recent SOTA approaches. The proposed WMamba achieves the highest AUC scores across all unseen datasets, demonstrating its exceptional generalization capability. This superior performance can be attributed to the synergistic effectiveness of DCConv and VMamba, which fully leverage wavelet information to facilitate the comprehensive extraction of forgery clues.

\subsubsection{Cross-Manipulation Evaluation}
In practice, defenders are often unaware of the exact forgery techniques used by attackers. Therefore, it is crucial to assess a model's ability to generalize across various types of manipulations. Following the protocol outlined in \cite{Shiohara_2022_CVPR}, we compare our method with approaches trained exclusively on real face videos from FF++. The results, shown in Table~\ref{cross-mani}, highlight WMamba's effectiveness in handling unseen manipulations.

\begin{table}[t]	
	\centering\renewcommand\arraystretch{1.2}\setlength{\tabcolsep}{8.1pt}
	\belowrulesep=0pt\aboverulesep=0pt
	\caption{Ablation study results for DCConv, showcasing its superiority over other deformable convolution paradigms. \label{DCConv_exp}}
	\begin{tabular}{c|cccc}
		\toprule
		\multirow{2}{*}{Method} & 
		\multicolumn{4}{c}{Test Set AUC (\%)} \\
		\cmidrule(lr){2-5}
		&\multicolumn{1}{c}{CDF} 
		&\multicolumn{1}{c}{DFDC} 
		&\multicolumn{1}{c}{DFDCP}
		&\multicolumn{1}{c}{FFIW} \\
		\midrule
		w/ DCN \cite{Dai_2017_ICCV} & 95.15 & 80.15 & \underline{87.77} & 85.53 \\
		w/ DCN-v2 \cite{Zhu_2019_CVPR} & 95.25 & 80.96 & 86.74 & 85.59 \\
		w/ DSConv \cite{Qi_2023_ICCV} & 94.42 & \underline{82.87} & 85.83 & \underline{86.26} \\
		w/o DCConv & \underline{96.06} & 80.24 & 87.21 & 85.88\\
		\midrule
		The Proposed & \textbf{96.29} & \textbf{82.97} & \textbf{89.62} & \textbf{86.59}\\ 
		\bottomrule
	\end{tabular}
\end{table}

\begin{table}[t]	
	\centering\renewcommand\arraystretch{1.2}\setlength{\tabcolsep}{3.6pt}
	\belowrulesep=0pt\aboverulesep=0pt
	\caption{Comparison of different VMamba backbones. \label{backbone}}
	\begin{tabular}{c|c|cccc|c}
		\toprule
		\multirow{2}{*}{Backbone} & 
		\multirow{2}{*}{Params} & 
		\multicolumn{5}{c}{Test Set AUC (\%)} \\
		\cmidrule(lr){3-7}
		&\multicolumn{1}{c|}{} 
		&\multicolumn{1}{c}{CDF} 
		&\multicolumn{1}{c}{DFDC} 
		&\multicolumn{1}{c}{DFDCP}
		&\multicolumn{1}{c|}{FFIW} 
		&\multicolumn{1}{c}{Avg.} \\
		\midrule
		VMamba-T & 30M & \underline{95.82} & \underline{82.00} & \underline{88.03} & 83.15 & 87.25 \\
		VMamba-S & 50M & \textbf{96.29} & \textbf{82.97} & \textbf{89.62} & \underline{86.59} & \textbf{88.87} \\
		VMamba-B & 89M & 95.29 & 81.64 & 88.09 & \textbf{87.29} & \underline{88.08} \\
		\bottomrule
	\end{tabular}
\end{table}

\subsection{Ablation Studies}
In this section, we present extensive ablation study results on HWFEB, DCConv, and VMamba. We also include visual demonstrations to highlight the efficacy of HWFEB and DCConv. Additional findings from our ablation studies are provided in Appendix~\ref{appendixc}.

\subsubsection{Structural Analysis of the HWFEB} 
We validate the design choices of HWFEB through comprehensive ablation studies conducted on five architectural variants. Specifically, our experiments investigate: (1) the impact of including the LL component in WFEMs (w/ LL), compared to our baseline using only LH, HL, and HH sub-bands for wavelet guidance; (2) the benefits of employing multi-scale wavelet representations over single-scale analysis (w/o Hierarchical); and (3) the effectiveness of the spatial gating mechanism versus alternative feature fusion strategies (w/ Addition, w/ Concatenation, and w/o Skip Connection). The cross-dataset evaluation results, presented in Table \ref{HWFEB}, validate our design decisions by demonstrating superior performance across all tested variants.

\subsubsection{Effectiveness of DCConv} 
To validate the efficacy of DCConv, we conduct comparative experiments against DCN \cite{Dai_2017_ICCV}, DCN-v2 \cite{Zhu_2019_CVPR}, DSConv \cite{Qi_2023_ICCV}, and a baseline network without DCConv. The cross-dataset evaluation results, shown in Table~\ref{DCConv_exp}, \textbf{clearly demonstrate that DCConv consistently outperforms other deformable convolution paradigms}, underscoring its superior capability to capture the slender structures of facial contours.

\begin{table}[t]	
	\centering\renewcommand\arraystretch{1.2}\setlength{\tabcolsep}{1.6pt}
	\belowrulesep=0pt\aboverulesep=0pt
	\caption{Performance and efficiency analysis of popular network architectures with and without HWFEB. ViT extracts features at a single scale, which makes it incompatible with HWFEB. We report the token numbers at the first stage of each network. Notably, VMamba outperforms both the CNN-based ConvNeXt and Transformer-based ViT and Swin. Additionally, VMamba showcases significantly better efficiency than ViT. Furthermore, integrating HWFEB consistently improves detection performance across all backbones. \label{others}}
	\begin{tabular}{c|c|c|c|cccc}
		\toprule
		\multirow{2}{*}{Method} & 
		\multirow{2}{*}{Params} & 
		\multirow{2}{*}{FLOPs} & 
		\multirow{2}{*}{Tokens} & 
		\multicolumn{4}{c}{Test Set AUC (\%)} \\
		\cmidrule(lr){5-8}
		&\multicolumn{1}{c|}{} 
		&\multicolumn{1}{c|}{} 
		&\multicolumn{1}{c|}{} 
		&\multicolumn{1}{c}{CDF} 
		&\multicolumn{1}{c}{DFDC} 
		&\multicolumn{1}{c}{DFDCP}
		&\multicolumn{1}{c}{FFIW} \\
		\midrule
		ConvNeXt-S & 50M & 8.7G & 3,136 & 94.39 & 80.23 & 86.26 & 82.04 \\
		+ HWFEB & 51M & 8.9G & 3,136 & 94.46 & \underline{82.17} & 86.22 & 84.60 \\
		\midrule
		ViT-B/16 & 86M & 17.6G & 197 & 93.64 & 79.66 & 85.73 & 83.92 \\
		\midrule
		Swin-S & 50M & 8.7G & 3,136 & 95.26 & 80.06 & 86.80 & 84.87 \\
		+ HWFEB & 51M & 8.9G & 3,136 & \underline{95.60} & 81.01 & \underline{87.91} & \underline{86.33} \\
		\midrule
		VMamba-S & 50M & 8.7G & 3,136 & {95.39} & {81.37} & {86.93} & {85.60}\\ 
		+ HWFEB & 51M & 8.9G & 3,136 & \textbf{96.29} & \textbf{82.97} & \textbf{89.62} & \textbf{86.59}\\ 
		\bottomrule
	\end{tabular}
\end{table}

\subsubsection{Analysis of VMamba} 
We systematically evaluate three versions of the VMamba backbone, namely VMamba-T (Tiny), VMamba-S (Small), and VMamba-B (Base). The cross-dataset assessment results, as presented in Table~\ref{backbone}, challenge the conventional assumption that larger architectures inherently deliver better performance. We conjecture that the greater capacity of VMamba-B leads to overfitting on the training set, consequently hindering its generalization to new, unseen examples. Notably, VMamba-S demonstrates superior performance compared to both its smaller and larger counterparts, establishing it as the optimal backbone for WMamba.

Furthermore, we compare VMamba against three widely used network architectures: the CNN-based ConvNeXt \cite{Liu_2022_CVPR} and the Transformer-based ViT \cite{dosovitskiy2020vit} and Swin \cite{Liu_2021_ICCV}. ViT processes features at a fixed $14\times 14$ spatial resolution, whereas other architectures adopt a hierarchical design, extracting features across multiple scales (up to $56\times 56$). Adapting ViT to support multi-scale feature extraction would drastically increase its computational cost by two orders of magnitude, making it impractical. Although Swin improves efficiency through windowed attention, this localized approach limits its ability to capture long-range spatial relationships. For a fair comparison, all backbones are pre-trained on ImageNet-1K and matched in parameter count. The cross-dataset evaluation results presented in Table~\ref{others} reveal three insights: \textbf{(1) VMamba outperforms both CNN- and Transformer-based architectures, highlighting its superior capability in face forgery detection; (2) VMamba achieves comparable computational efficiency to ConvNeXt and Swin in terms of FLOPs and token counts, while significantly surpassing ViT in this regard; and (3) incorporating the HWFEB consistently enhances detection performance, demonstrating its effectiveness as a plug-and-play module.}

\begin{figure}[t]
	\begin{center}
		\begin{minipage}{1\linewidth}
			{\includegraphics[width=0.94\linewidth]{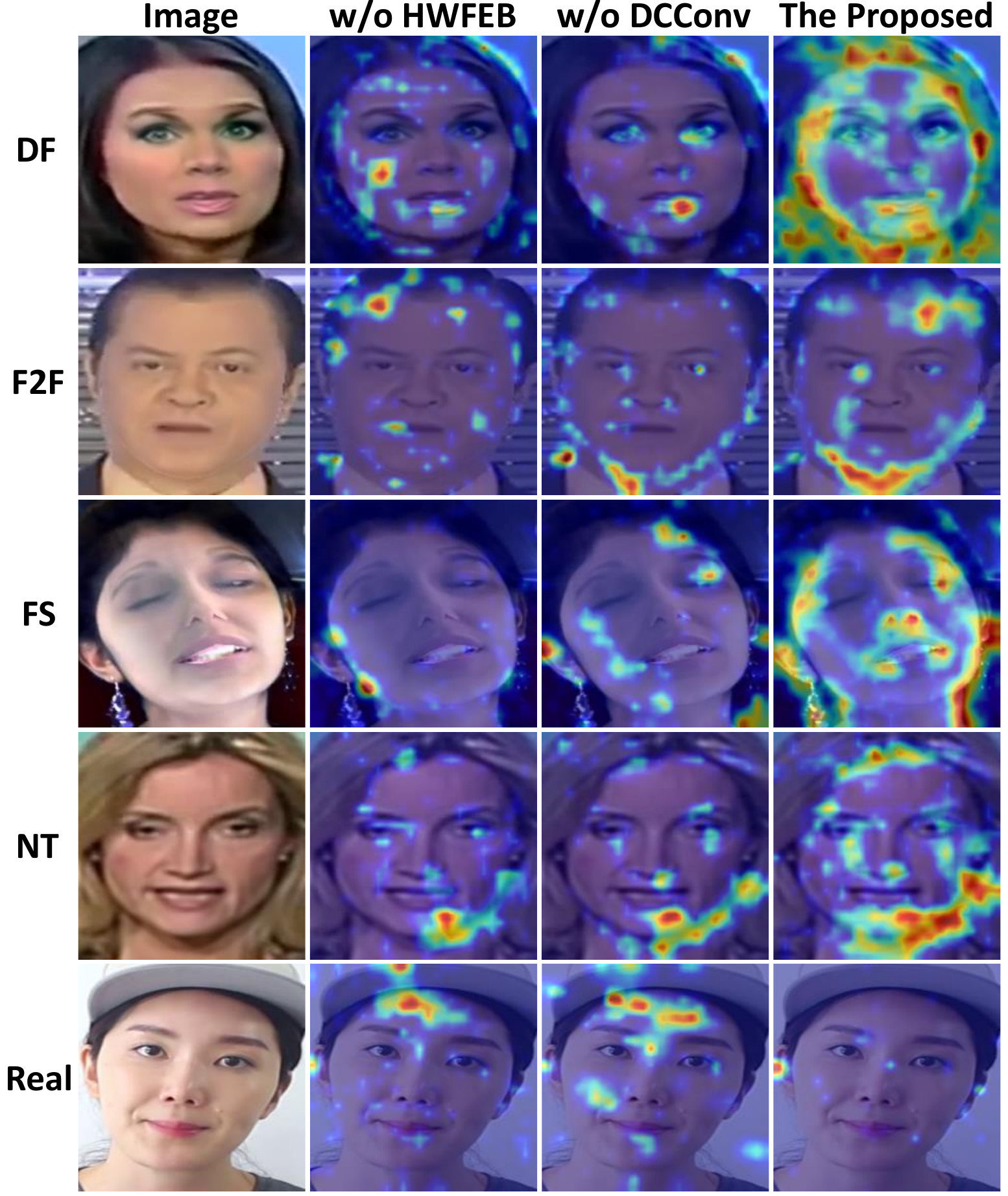}}
			\centering
		\end{minipage}
	\end{center}
	\caption{Saliency maps on four distinct manipulated samples and one real sample from the FF++ dataset. For the forged instances, HWFEB and DCConv guide our model's attention to key facial contours that are rich in forgery artifacts. \label{grad_cam}}
\end{figure}

\subsubsection{Visual Analysis of HWFEB \& DCConv} 
We further validate the effectiveness of HWFEB and DCConv using Grad-CAM visualizations \cite{Selvaraju_2017_ICCV}, which highlight the spatial regions most critical to the model's decision-making process. The saliency maps in Figure~\ref{grad_cam} indicate that, for fake samples, HWFEB and DCConv direct the model's attention to key facial contours rich in forgery artifacts. In contrast, for the real sample, our model exhibits minimal activation.

\section{Conclusion}
In this paper, we propose WMamba, a wavelet-based feature extractor built upon the Mamba architecture. WMamba fully considers the slender, fine-grained, and globally distributed nature of facial contours, maximizing the utility of wavelet data via two key innovations. First, we introduce DCConv, which employs a learnable coordinate axis to capture a broader range of slender structures. Second, by leveraging the Mamba architecture, which excels at modeling long-range spatial dependencies with linear complexity, WMamba effectively captures fine-grained, globally distributed forgery clues from small image patches. Our method achieves SOTA performance in both cross-dataset and cross-manipulation evaluations, demonstrating its effectiveness in face forgery detection.

\begin{acks}
This work was supported in part by Chinese National Natural Science Foundation Projects U23B2054, 62276254, 62306313, the Beijing Science and Technology Plan Project Z231100005923033, Beijing Natural Science Foundation L221013, the Science and Technology Development Fund of Macau Project 0140/2024/AGJ, and InnoHK program.
\end{acks}

\bibliographystyle{ACM-Reference-Format}
\balance
\bibliography{sample-base}

\clearpage
\appendix

\section{Comparison Between DSConv \& DCConv}
\label{appendixa}
\begin{figure}[t]
	\begin{center}
		\begin{minipage}{1\linewidth}
			{\includegraphics[width=1\linewidth]{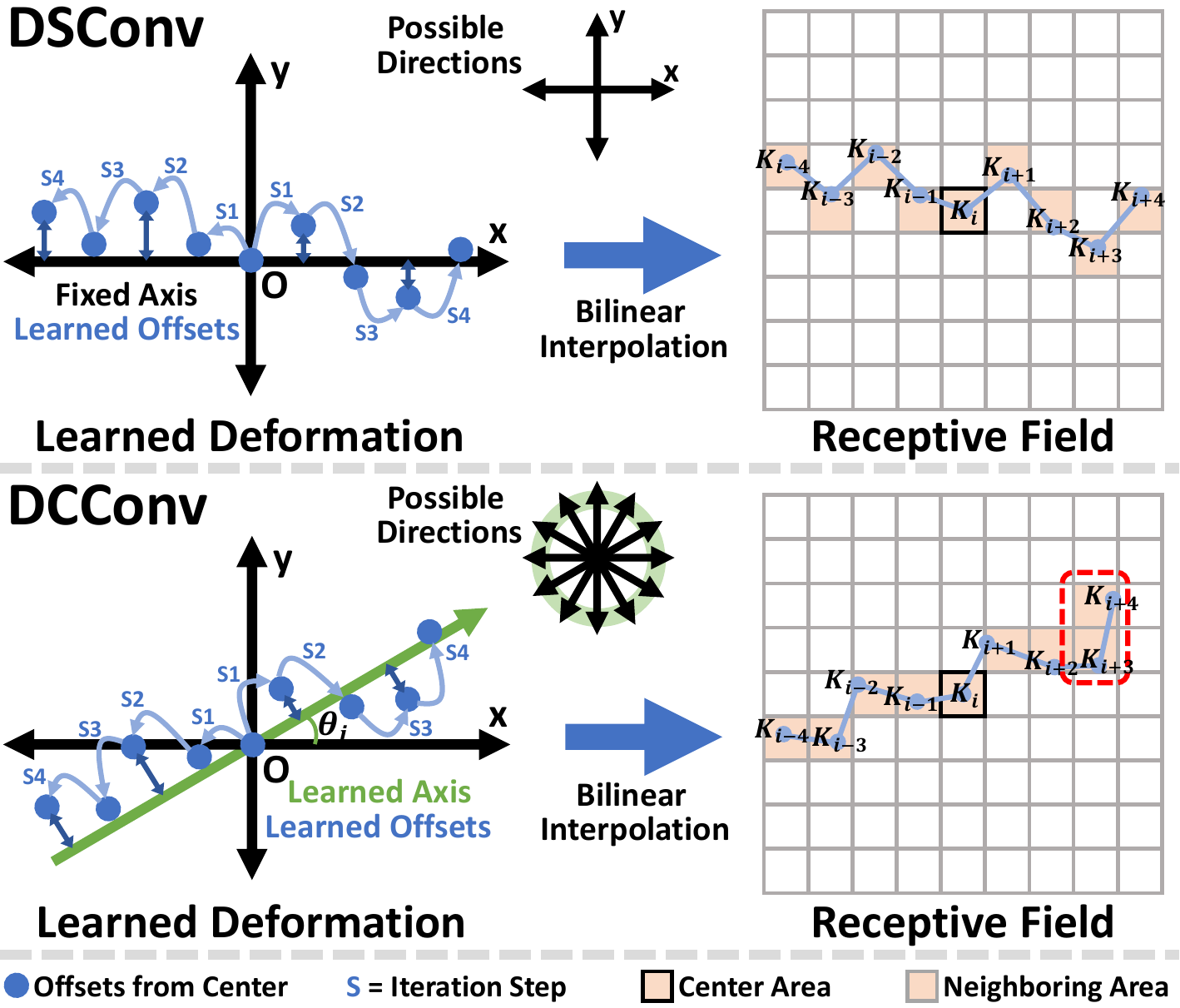}}
			\centering
		\end{minipage}
	\end{center}
	\caption{Visual comparison between DSConv and the proposed DCConv, initialized along the x-axis. DSConv learns perpendicular offsets to model slender structures oriented along a predefined direction. In contrast, DCConv predicts both perpendicular offsets and axis orientation, enabling it to represent slender geometries aligned in any direction. This added flexibility allows DCConv to capture a broader range of slender structures, including the shape highlighted in the red box, which DSConv is unable to model. \label{ds_dc}}
\end{figure}

Figure~\ref{ds_dc} visually compares DSConv with the proposed DCConv. DSConv captures slender structures by learning offsets aligned strictly along the x-axis or y-axis. In contrast, DCConv simultaneously predicts both the offsets and the axis orientation, allowing it to represent slender geometries in any direction. As a result, DCConv is capable of modeling a broader range of slender structures.

\section{Implementation Details}
\label{appendixb}
This study uses the synthetic data generation framework, preprocessing techniques, and data augmentation strategies in SBI \cite{Shiohara_2022_CVPR}.

\subsection{Synthetic Data Generation.} 
The SBI framework generates fake face images from authentic face samples by mimicking the deepfake generation process. Specifically, a real image is first input into a Source-Target Generator (STG) and a Mask Generator (MG). The STG applies a series of image transformations (detailed later) to create pseudo source and target images, while the MG produces a blending mask based on pre-detected facial landmarks and further augments it to enhance diversity. Finally, the pseudo source and target images are blended using the generated mask, resulting in a pseudo-fake face sample.

\subsection{Preprocessing.} 
The 81-point facial landmark shape predictor from Dlib \cite{king2009dlib} is utilized to extract facial landmarks, which are required exclusively during the training phase. Additionally, the RetinaFace \cite{Deng_2020_CVPR} is employed to generate facial bounding boxes. During training, the bounding box is cropped with a random margin ranging from 4\% to 20\%, while for inference, a fixed margin of 12.5\% is applied.

\subsection{Data Augmentation.} 
The image processing toolbox presented in \cite{buslaev2020albumentations} is employed for data augmentation. Within the STG of SBI, image transformations such as RGBShift, HueSaturationValue, RandomBrightnessContrast, Downscale, and Sharpen are applied to create pseudo source and target images. During the training phase, real samples undergo augmentations such as ImageCompression, RGBShift, HueSaturationValue, and RandomBrightnessContrast. These operations greatly improve the generalization capability of our model.

\subsection{Other Details.} 
We sample eight frames from each video during training and 32 frames per video for inference. If a frame contains multiple detected faces, the classifier is applied to each face individually, and the highest fakeness score among them is selected as the predicted confidence for that frame. Additionally, the input images are resized to a spatial resolution of $224 \times 224$ during both training and testing.

\begin{figure*}[t]
	\begin{center}
		\begin{minipage}{1\linewidth}
			{\includegraphics[width=1\linewidth]{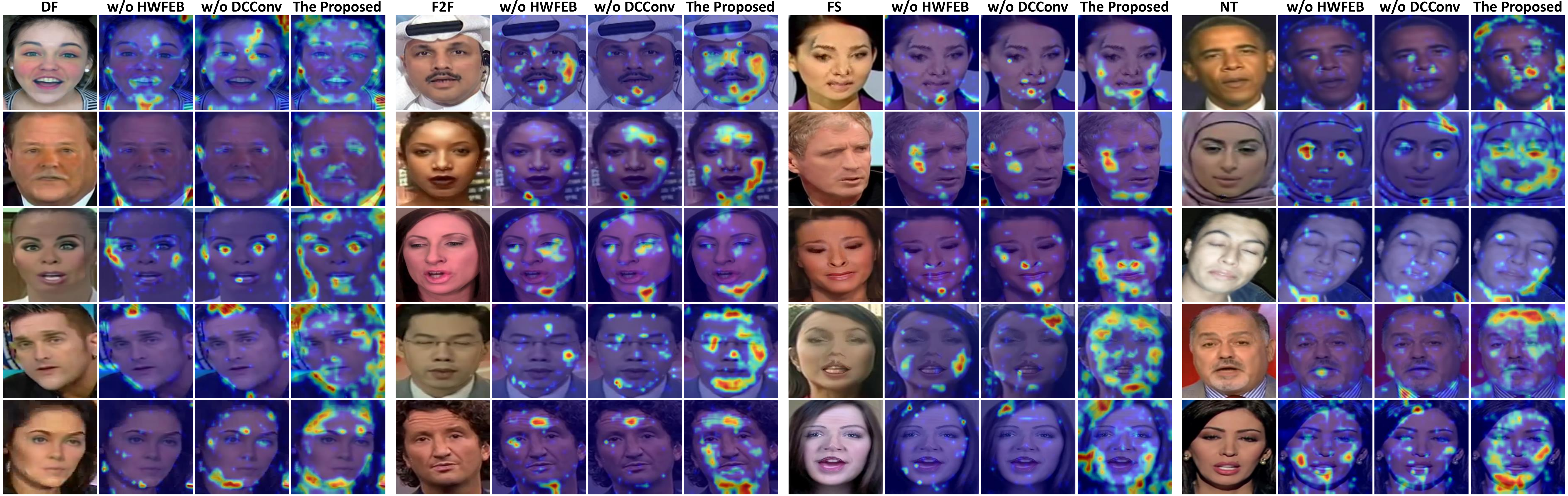}}
			\centering
		\end{minipage}
	\end{center}
	\caption{Saliency maps for a variety of fake samples from the FF++ dataset, where redder regions indicate areas of higher model attention. The HWFEB and DCConv effectively direct our model's attention toward critical facial contours. \label{grad}}
\end{figure*}

\begin{table}[h]	
	\centering\renewcommand\arraystretch{1.2}\setlength{\tabcolsep}{6.0pt}
	\belowrulesep=0pt\aboverulesep=0pt
	\caption{Cross-manipulation evaluation results for the HWFEB and DCConv. These two components effectively enhance the overall performance of our model. \label{supp1}}
	\begin{tabular}{c|cccc|c}
		\toprule
		\multirow{2}{*}{Method} & 
		\multicolumn{5}{c}{Test Set AUC (\%)} \\
		\cmidrule(lr){2-6}
		&\multicolumn{1}{c}{DF} 
		&\multicolumn{1}{c}{F2F} 
		&\multicolumn{1}{c}{FS}
		&\multicolumn{1}{c|}{NT}
		&\multicolumn{1}{c}{FF++} \\
		\midrule
		w/o HWFEB & \textbf{100} & 99.93 & {99.87} & 98.38 & 99.55 \\
		w/o DCConv & {99.99} & {99.97} & {99.87} & {98.83} & {99.67} \\
		\midrule
		The Proposed & \textbf{100} & \textbf{99.98} & \textbf{99.94} & \textbf{98.88} & \textbf{99.70}\\      
		\bottomrule
	\end{tabular}
\end{table}

\begin{table}[h]
	\centering\renewcommand\arraystretch{1.2}\setlength{\tabcolsep}{4.5pt}
	\belowrulesep=0pt\aboverulesep=0pt
	\caption{Ablation study results on different HWFEB stages. \label{stage_ablation}}
	\begin{tabular}{cccc|cccc|c}
		\toprule
		\multicolumn{4}{c|}{Stages} & 
		\multicolumn{5}{c}{Test Set AUC (\%)} \\
		\cmidrule(lr){1-4}\cmidrule(lr){5-9}
		\multicolumn{1}{c}{1} 
		&\multicolumn{1}{c}{2} 
		&\multicolumn{1}{c}{3} 
		&\multicolumn{1}{c|}{4}
		&\multicolumn{1}{c}{CDF} 
		&\multicolumn{1}{c}{DFDC} 
		&\multicolumn{1}{c}{DFDCP} 
		&\multicolumn{1}{c|}{FFIW}
		&\multicolumn{1}{c}{Avg.} \\
		\midrule
		\checkmark &  &  &  & 94.76 & 81.79 & 87.88 & 85.03 & 87.34 \\
		\checkmark & \checkmark &  &  & 95.07 & 81.66 & 88.17 & 84.63 & 87.38 \\
		\checkmark & \checkmark & \checkmark &  & 95.99 & 82.79 & 89.44 & \textbf{86.95} & 88.79 \\
		\checkmark & \checkmark & \checkmark & \checkmark & \textbf{96.29} & \textbf{82.97} & \textbf{89.62} & 86.59 & \textbf{88.87} \\
		\bottomrule
	\end{tabular}
\end{table}

\section{Experimental Results}
\label{appendixc}
In this section, we present additional experimental results that are omitted from the main text due to space constraints.

\begin{table}[t]	
	\centering\renewcommand\arraystretch{1.2}\setlength{\tabcolsep}{3.2pt}
	\belowrulesep=0pt\aboverulesep=0pt
	\caption{Ablation study results for scanning methods. \label{scan}}
	\begin{tabular}{c|cccc}
		\toprule
		\multirow{2}{*}{Method} & 
		\multicolumn{4}{c}{Test Set AUC (\%)} \\
		\cmidrule(lr){2-5}
		&\multicolumn{1}{c}{CDF} 
		&\multicolumn{1}{c}{DFDC} 
		&\multicolumn{1}{c}{DFDCP}
		&\multicolumn{1}{c}{FFIW}\\
		\midrule
		Single-directional Scan \cite{gu2023mamba} & 94.37 & 80.39 & 87.61 & 85.98 \\
		Bidirectional Scan \cite{10.5555/3692070.3694654} & 95.40 & 81.32 & 88.42 & 86.47 \\
		\midrule
		Cross-Scan \cite{liu2024vmamba} & \textbf{96.29} & \textbf{82.97} & \textbf{89.62} & \textbf{86.59} \\
		\bottomrule
	\end{tabular}
\end{table}

\subsection{Additional Analysis of HWFEB \& DCConv.}
The HWFEB integrates wavelet information into the VMamba model, while the DCConv effectively captures the slender structures of facial contours. As visually demonstrated in Figure~\ref{grad_cam}, these components are instrumental in directing the model's attention toward key facial contours, which are often rich in forgery artifacts. To further substantiate this observation, we present additional experimental results. As illustrated in Figure~\ref{grad}, incorporating HWFEB and DCConv enhances the model's ability to accurately focus on essential facial contours during decision-making. Moreover, the cross-manipulation evaluation results in Table~\ref{supp1} highlight the contributions of these two components to the model's performance. In Table~\ref{HWFEB}, we assess the impact of applying DWT only at the first stage of the backbone. Here, we extend this analysis to examine the individual contributions of each HWFEB stage. The cross-dataset evaluation results in Table~\ref{stage_ablation} indicate that all stages work together to enhance performance, validating our hierarchical design.

\subsection{Additional Analysis of VMamba.}
\begin{figure}[t]
	\begin{center}
		\begin{minipage}{1\linewidth}
			\begin{minipage}[t]{0.49\linewidth}
				{\includegraphics[width=1\linewidth]{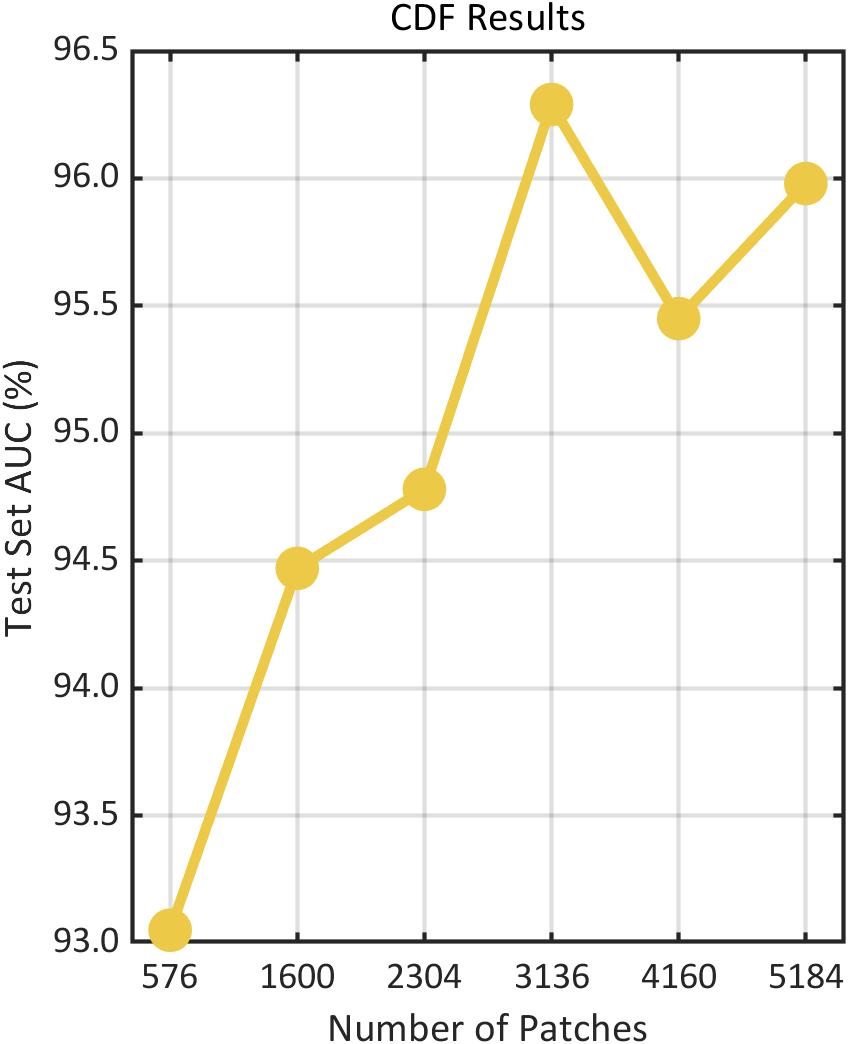}}
				\centering
			\end{minipage}
			\begin{minipage}[t]{0.49\linewidth}
				{\includegraphics[width=1\linewidth]{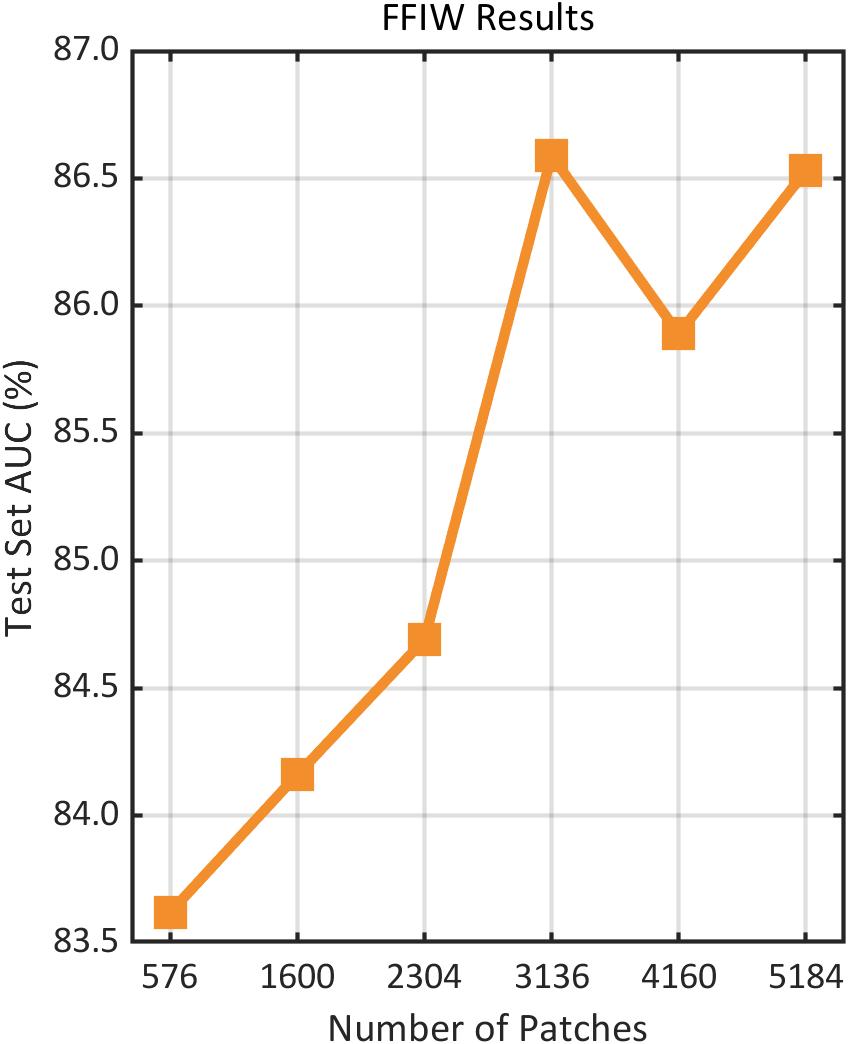}}
				\centering
			\end{minipage}
		\end{minipage}
	\end{center}
	\caption{Effect of patch count on model performance. The horizontal axis shows the number of patches following the Stem module in the VMamba model. Overall, performance improves with an increase in patch count. Notably, the peak performance is observed at a patch count of 3136, which corresponds to an input size of $224\times 224$ pixels, the same resolution utilized during the pretraining of VMamba-S. \label{tokens}}
\end{figure}

We investigate the effect of patch count on model performance by systematically varying the input image sizes. The cross-dataset evaluation results in Figure~\ref{tokens} reveal that, in general, model performance improves as the number of patches increases. The peak performance is observed at a patch count of 3136, which corresponds to an input size of $224 \times 224$ pixels, the same resolution used during the pretraining of the VMamba-S model. Based on these findings, we choose to maintain the pretraining resolution rather than arbitrarily increasing the input size. Additionally, we compare the cross-scan used in VMamba with single-directional \cite{gu2023mamba} and bidirectional \cite{10.5555/3692070.3694654} scans. The cross-dataset evaluation results in Table~\ref{scan} support the validity of our design choice.

\subsection{Analysis of Kernel Length for DCConv.}
We systematically evaluate four different kernel lengths for DCConv. Based on the cross-dataset assessment results in Table~\ref{kernel_length}, a kernel length of 9 is selected as it delivers the best performance.

\subsection{Model Robustness.}
To assess the robustness of our method, we conduct experiments across four real-world degradation conditions: compression, occlusion, noise, and blur. For compression, we apply the JPEG standard with quality settings of 95 (low), 85 (medium), and 75 (high). For occlusion, we randomly mask 2\% (low), 5\% (medium), and 10\% (high) of the image pixels. For noise, we introduce Gaussian noise with standard deviations of 0.01 (low), 0.05 (medium), and 0.1 (high). For blur, we use Gaussian blur with kernel/sigma values of 5/1 (low), 7/2 (medium), and 9/3 (high). The cross-dataset results on the CDF dataset, shown in Table~\ref{degrad}, demonstrate that WMamba consistently and significantly outperforms SBI across all degradation conditions, with particularly strong performance in occlusion scenarios.

\begin{table}[t]	
	\centering\renewcommand\arraystretch{1.2}\setlength{\tabcolsep}{6.9pt}
	\belowrulesep=0pt\aboverulesep=0pt
	\caption{Ablation study results on DCConv's length. \label{kernel_length}}
	\begin{tabular}{c|cccc|c}
		\toprule
		\multirow{2}{*}{Length} & 
		\multicolumn{5}{c}{Test Set AUC (\%)} \\
		\cmidrule(lr){2-6}
		&\multicolumn{1}{c}{CDF} 
		&\multicolumn{1}{c}{DFDC} 
		&\multicolumn{1}{c}{DFDCP}
		&\multicolumn{1}{c|}{FFIW}
		&\multicolumn{1}{c}{Avg.} \\
		\midrule
		5 & 95.13 & 82.67 & {87.10} & 86.10 & 87.75 \\
		7 & 95.66 & \textbf{82.99} & 88.94 & \textbf{86.97} & {88.64} \\
		9 & \textbf{96.29} & {82.97} & \textbf{89.62} & {86.59} & \textbf{88.87} \\
		11 & {95.70} & 81.89 & {89.18} & 86.06 & 88.21 \\
		\bottomrule
	\end{tabular}
\end{table}

\begin{table}[t]
	\centering
	\renewcommand\arraystretch{1.2}\setlength{\tabcolsep}{5.9pt}
	\belowrulesep=0pt\aboverulesep=0pt
	\caption{Model robustness across various types and severities of image degradation. We compare WMamba with SBI \cite{Shiohara_2022_CVPR}. \label{degrad}}
	\begin{tabular}{c|c|ccc}
		\toprule
		\multirow{2}{*}{{Degradation}} & \multirow{2}{*}{{Method}} & \multicolumn{3}{c}{$\Delta$AUC by Severity (\%)} \\
		\cmidrule(lr){3-5}
		& & {Low} & {Medium} & {High} \\
		\midrule
		\multirow{2}{*}{{Compression}} 
		& {SBI \cite{Shiohara_2022_CVPR}} & -0.73 & -1.66 & -2.29 \\
		& {WMamba} & \textbf{-0.52} & \textbf{-1.22} & \textbf{-2.13} \\
		\midrule
		\multirow{2}{*}{{Occlusion}} 
		& {SBI \cite{Shiohara_2022_CVPR}} & -24.71 & -32.31 & -36.08 \\
		& {WMamba} & \textbf{-12.87} & \textbf{-22.71} & \textbf{-28.15} \\
		\midrule
		\multirow{2}{*}{{Noise}} 
		& {SBI \cite{Shiohara_2022_CVPR}} & -14.57 & -29.75 & -31.33 \\
		& {WMamba} & \textbf{-7.20} & \textbf{-27.16} & \textbf{-29.01} \\
		\midrule
		\multirow{2}{*}{{Blur}} 
		& {SBI \cite{Shiohara_2022_CVPR}} & -3.20 & -12.77 & -20.30 \\
		& {WMamba} & \textbf{-2.51} & \textbf{-9.51} & \textbf{-16.97} \\
		\bottomrule
	\end{tabular}
\end{table}

\begin{table}[t]	
	\centering\renewcommand\arraystretch{1.2}\setlength{\tabcolsep}{10.5pt}
	\belowrulesep=0pt\aboverulesep=0pt
	\caption{Quantitative evaluation results on the DRIVE dataset. The proposed DCConv achieves the best results. \label{drive}}
	\begin{tabular}{c|cc|c}
		\toprule
		\multirow{2}{*}{Method} & 
		\multicolumn{3}{c}{Evaluation Metric (\%)} \\
		\cmidrule(lr){2-4}
		&\multicolumn{1}{c}{IoU} 
		&\multicolumn{1}{c|}{Dice}
		&\multicolumn{1}{c}{Avg.}  \\
		\midrule
		U-Net \cite{ronneberger2015u} & 66.57 & 79.90 & 73.24\\
		U-Net + DSConv \cite{Qi_2023_ICCV} & {67.70} & {80.69} & {74.20} \\
		\midrule
		U-Net + DCConv & \textbf{68.70} & \textbf{81.41} & \textbf{75.06} \\ 
		\bottomrule
	\end{tabular}
\end{table}

\begin{figure}[h]
	\begin{center}
		\begin{minipage}{1\linewidth}
			{\includegraphics[width=1\linewidth]{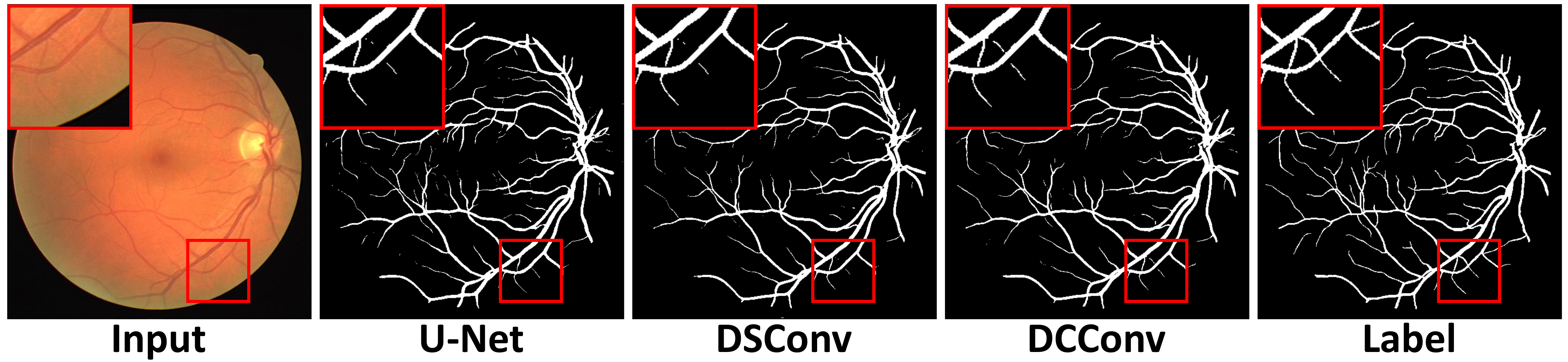}}
			\centering
		\end{minipage}
	\end{center}
	\caption{Qualitative evaluation results on the DRIVE dataset. Notably, DCConv better segments subtle vessels. \label{seg}}
\end{figure}

\subsection{The Versatility of DCConv.}
We assess the versatility of DCConv using the Digital Retinal Images for Vessel Extraction (DRIVE) dataset \cite{1282003}, a widely recognized benchmark for medical image segmentation. Notably, the segmentation targets in the DRIVE dataset exhibit slender structures, making it a suitable testbed for our approach. The dataset comprises 20 training and 20 testing images, each with dimensions of $565 \times 584 \times 3$. To facilitate training, we crop the original training samples into 4,000 smaller patches, each of size $64\times 64\times 3$. For our experiments, we integrate DCConv into the U-Net architecture \cite{ronneberger2015u}, replacing its standard convolutional layers. We then compare the performance of DCConv-enhanced U-Net against two baselines: the original U-Net and a variant where the convolutional layers are replaced by DSConv \cite{Qi_2023_ICCV}. To ensure a fair comparison, all methods are trained under the same experimental conditions. The training process leverages a combined objective function that integrates Dice Loss \cite{milletari2016v} and Focal Loss \cite{Lin_2017_ICCV}. Each model is trained for 50 epochs with a batch size of 16 using the Adam optimizer. The initial learning rate is set to 1e-3 and is reduced by half every 10 epochs. The quantitative evaluation results presented in Table \ref{drive}, assessed using the Intersection over Union (IoU) and Dice metrics, demonstrate that our method achieves superior performance compared to others. Additionally, the qualitative evaluation results in Figure~\ref{seg} highlight DCConv's exceptional capability to capture subtle and slender structures. Overall, these findings underscore the effectiveness of DCConv as a versatile, plug-and-play module for improving the representation of slender structures.


\end{document}